\documentclass[a4paper,twocolumn,twoside,10pt]{ranlp}

\usepackage{booktabs}

\usepackage[usenames]{color}

\usepackage{enumerate}
\usepackage{latexsym}
\usepackage{url}
\usepackage{lingmacros}
\usepackage{graphicx}
\newcommand{\tlfi}{{\sc TLFi}}

\begin{document}

\title{\textbf{Grouping synonyms by definitions}}

\author{
Ingrid Falk\thanks{The research presented in this paper was partially supported by the
TALC theme of the CPER "Mod\'elisation, Information et Syst\`emes
Num\'eriques" funded by the R\'{e}gion Lorraine. We also gratefully
acknowledge the ATILF for letting us access their synonym database
and the TLFi.
}\\
INRIA / Universit\'{e} Nancy 2\\
\email{ingrid.falk}{loria.fr}\\
\and
Claire Gardent\\
CNRS / LORIA, Nancy\\
\email{claire.gardent}{loria.fr}\\
\and 
Evelyne Jacquey\\
CNRS / ATILF, Nancy\\
\email{evelyne.jacquey}{atilf.fr}\\
\and 
Fabienne Venant\\
Universit\'e Nancy 2 / LORIA, Nancy\\
\email{fabienne.venant}{loria.fr}
}
%
%
\date{} 
\maketitle 

\thispagestyle{empty}	     \pagestyle{empty}

\begin{abstract}
  We present a method for grouping the synonyms of a lemma according
  to its dictionary senses. The senses are defined by a large machine
  readable dictionary for French, the \tlfi\ (Tr\'esor de la langue
  fran\c caise informatis\'e) and the synonyms are given by 5 synonym
  dictionaries (also for French). To evaluate the proposed method, we
  manually constructed a gold standard where for each (word,
  definition) pair and given the set of synonyms defined for that word
  by the 5 synonym dictionaries, 4 lexicographers specified the set of
  synonyms they judge adequate. While inter-annotator agreement ranges
  on that task from 67\% to at best 88\% depending on the annotator
  pair and on the synonym dictionary being considered, the automatic
  procedure we propose scores a precision of 67\% and a recall of
  71\%. The proposed method is compared with related work namely,
  word sense disambiguation, synonym lexicon acquisition and WordNet
  construction.
\end{abstract}



\section{Introduction}
\label{sec:introduction}

Synonymic resources for French are still limited in scope, quality
and/or availability.  Thus the French WordNet (\textsc{Frewn}) created within
the EuroWordNet project \cite{jacquin07} has limited scope (3 777 verbs and 14 618 nouns vs. 7 384 verbs and 42 849 nouns in the morphological lexicon for French Morphalou) 
and has not been widely used
mainly due to licensing issues. The alternative open-source WordNet
for French called \textsc{Wolf} (WordNet Libre du Francais, \cite{sagot08a})
remedies the first shortcoming (restrictive licensing) and aims to
achieve a wider coverage by automating the WordNet construction
process using an {\it extend} approach which in essence, translates
the synsets from Princeton WordNet (PWN) into French. 
However, compared to 
Morphalou,
\textsc{Wolf} is still incomplete (979 verbs and 34 827 nouns). Finally, the synonym lexicon DicoSyn
\cite{dicosyn04} is restricted to assigning sets of synonyms to lemmas
thereby lacking both categorial information and definitions. 

In this paper, we present a method for grouping synonyms by senses and
evaluate it on the synonyms given by 5 synonym dictionaries included
in the \textsc{Atilf} synonym database. The long term aim is to apply this
method to these synonym dictionaries so as to build a uniform
synonymic resource for French in which each lemma is assigned a part
of speech, a set of (\tlfi ) definitions and for each given
definition, a set of synonyms. The resulting resource should
complement DicoSyn and \textsc{Wolf}. Contrary to DicoSyn, it will include
categorial information and associate groups of synonyms with
definitions. It will furthermore complement \textsc{Wolf} by providing an
alternative synonymic resource which, being built on handbuilt high
quality resources, should differ from \textsc{Wolf} both in coverage and in
granularity.

The paper is structured as follows. Section~\ref{sec:data} presents
the data we are working from, namely a set of synonym dictionaries for
French and the \tlfi , the largest machine readable dictionary
available for French. Section~\ref{sec:method} describes the basic
algorithm used to assign a verb synonym to a given definition. Section~\ref{sec:experiments} presents the experiments we did to assess the
impact of the similarity measures used and of a linguistic
preprocessing on the definitions. Section~\ref{sec:related} discusses
related work. Section~\ref{sec:conclusion} concludes and gives
pointers for further research.

\section{The source data}
\label{sec:data}

We have at our disposal a general purpose machine readable dictionary
for French, the Tr\'esor de la Langue Fran\c caise informatis\'e
(\tlfi , \cite{tlfi,tlf}) and 5 synonym dictionaries namely, {\it
  Dictionnaire des synonymes de la langue fran\c{c}aise} \cite{syndic1}, {\it Dictionnaire des
  synonymes } \cite{syndic2}, {\it Nouveau dictionnaire des synonymes
}
\cite{syndic3}, {\it Dictionnaire alphab\'etique et analogique de la
  langue fran\c caise} \cite{syndic5}, {\it Grand Larousse 
  de la Langue Fran\c{c}aise} \cite{syndic4}. 

One driving motivation behind our  method is the question of how to
merge these 5 synonym lexicons in a meaningful way. Indeed although
one of them (namely, \cite{syndic5}) covers most of the verbs
present in the five synonym lexicons (5 027 verbs out of 5 736), a
merge of the lexicons would permit an increased ``synonymic coverage''
(11 synonyms in average per verb with the 5 lexicons against 6 per
verb using only \cite{syndic5}). To merge the five lexicons, we plan
to apply the  method presented here to each of the synonyms
assigned to a word by the 5 synonym lexicons. In this way, we aim to 
obtain a merged lexicon in which each word is associated with a part
of speech, a set of \tlfi\ definitions and for each definition, the
set of synonyms associated to this definition.

For our experiment, we worked on a restricted dataset. First, we
handled only verbs. Since they are in average more
polysemous\footnote{The average polysemy recorded by the Princeton
  WordNet for the various parts of speech is: 2.17 for verbs, 1.4 for
  adjectives, 1.25 for adverbs and 1.24 for nouns.}  than other
categories, they nevertheless provide an interesting
benchmark. Second, we based our evaluation on a single synonym
dictionary, namely \cite{syndic5}. As mentioned above, this is the
largest of the five lexicons (cf. Fig.~\ref{tab:verbs-tlf-and-synodict}). Moreover, it is unlikely that the
quality of the results obtained vary greatly when considering more
synonyms since, as we shall see in Section~\ref{sec:method}, the
synonym-to-definition mapping performed by our method is independent
of the number of synonyms assigned to a given word.
\paragraph{The \tlfi  }is the largest machine readable dictionary available for
French (54 280 entries, 92 997 lemmas, 271 166 definitions, 430 000
examples). It has a rich XML markup which supports a selective
treatment of entry subfields. Moreover, the definitions have been
part-of-speech tagged and lemmatised.

For our experiment, we extracted from the \tlfi\ all the verb entries
and their associated definitions. Definitions were extracted by
selecting the XML elements identifying an entry definition and
checking their content. If a selected definitional element contained
either some text (i.e., a definition), a synonym or a domain
specification, the XML element was taken to indeed identify a
definition. Else, no definition was stored. In this way, XML elements
that did not contain any definitional information such as
subdefinitions containing only examples, were not taken into account.

For each selected definitional element, a definition index was then
constructed by taking the open class lemmas associated with the
definition and, if any, the synonyms and/or the domain information
contained in the definitional element. For instance, given the
\tlfi\ definitions for {\it projeter} ({\it to project}) listed in
Fig.~\ref{ex:defs}, the indexes extracted will be as indicated below each
definition. In (\ref{item:def_a}), the index contains the open class
lemmas of the definition; in (\ref{item:def_b}) the domain information
is also included and in (\ref{item:def_c}), synonymic information is
added.
\begin{figure}
  \centering
  \begin{enumerate}[a.]
  \item\label{item:def_a} Jeter loin en avant avec force.\\
    {\scriptsize\emph{To throw far ahead and with strength}}\\
    \emph{$\langle$ jeter, loin, avant, force $\rangle$}
  \item\label{item:def_b} {\it \textsc{cin. audiovisuel}}. Passer dans un projecteur.\\
    {\scriptsize\emph{\textsc{cin. audiovisual}}. To show on a projector.}\\
    \emph{$\langle$ cin\'ema, audiovisuel, passer, projecteur $\rangle$}
  \item\label{item:def_c} Eclaircir. Synon. jeter quelque lumi\` ere\\
    {\scriptsize\emph{To lighten. Synonym. To throw some light.}}\\
    \emph{$\langle$ lighten. throw, light $\rangle$}
  \end{enumerate}
  \caption{Some definitions and extracted indexes for \emph{projeter} (\emph{to project}).} 
  \label{ex:defs}
\end{figure}
%
\paragraph{The synonym dictionaries.} 
 The table in Fig.~\ref{tab:verbs-tlf-and-synodict} gives a quantitative summary
 of the data contained in the five available synonym
 dictionaries. Each entry in the synonym dictionaries is associated
 with one or more sets of synonyms, each set corresponding to a
 different meaning of the entry. The synonym dictionaries however
 contain neither part of speech information nor definitions. An
 example entry of \cite{syndic5} is given in Figure~\ref{fig:abandonner}.  For the experiment, we extracted the verb
 entries (using a morphological lexicon) of these dictionaries that
 were also present in the \tlfi . Synonyms or entries that were
 present in the synonym dictionaries but not in the \tlfi\ were
 discarded.
\begin{figure}[h]
  \small
  \centering
  \begin{tabular}[h]{|lcccr@{.}l|}
    \toprule
    \textbf{Syn. Dic.} & Verbs & -Refl & +Refl & \multicolumn{2}{c|}{Syn/verb}\\
    \midrule
    \textbf{Bailly}           & 2600 & 2370 & 230 & 1& \\
    \textbf{Benac}            & 2656 & 2298 & 358 & 1&5 \\
    \textbf{Chazaud}       & 3808 & 3267 & 541 & 5&25 \\
    \textbf{Larousse}         & 3835 & 3194 & 641 & 4&7 \\
    \textbf{Rey}  & 5027 & 4071 & 956 & 6&\\
    \midrule
    \textbf{ALL} & 5736 & 4554 & 1182 & 11& \\
    \bottomrule
  \end{tabular}
  \caption{Verbs from \textsc{tlf}i, also present in the synonym
    dictionaries. -Refl indicates the number of non reflexive verb
    entries ({\it laver}), +Refl the number of reflexive verb entries
    ({\it se laver}). }
  \label{tab:verbs-tlf-and-synodict}
\end{figure}
\paragraph{Reference.} To evaluate our results, we built a reference sample as follows. 
First, we selected a sample of French verbs using the combination of
three features: genericity, polysemy and frequency. Each feature could
have one of the three values ``high'', ``medium'' and ``low'' thus
yielding a sample of 27 verbs. Genericity was assessed using the
position of the verb in the French EuroWordNet (the higher the more
generic). Polysemy was defined by the number of definitions assigned
to the verb by the \tlfi. Frequency was extracted from a frequency
list built from 10 years of Le Monde newspaper parsed with the Syntex
parser \cite{syntex}.

For these 27 verbs, we extracted the corresponding definitions and
synonyms from the \tlfi\ and the synonym dictionaries respectively. To
facilitate the assignment by the annotators of synonyms to
definitions, we manually reconstructed some of the definitions from
the information contained in the \tlfi\ entries. Indeed a dictionary
entry has a hierarchical structure (a definition can be the child of
another definition) which is often used by the lexicographer to omit
information in definitions occurring lower down in the hierarchy. The
assumption is that the missing information is inherited from the
higher levels. To facilitate the assignment by the annotator of a
given synonym to a given definition, we manually reconstructed the
information that had been omitted on an inheritance assumption. Note
though that this manual reconstruction is only intended to facilitate
the annotation task. It does not affect the evaluation since the
numbering of the definitions within a given dictionary entry remains
the same and what is being compared is solely the assignment of
synonyms to definition identifiers made by the system and that made by
the annotators. 

Third, we asked four professional lexicographers to manually assign
synonyms to definitions. The lexicographers were given for each verb
$v$ in the sample, the set of (possibly reconstructed) definitions
assigned by the \tlfi\ to $v$ and the set of synonyms associated to $v$
by the synonym base. They then had to decide which
definition(s) the synonym should be associated with.
%

We computed the agreement rate between pairs of annotators and all
four annotators. No pair achieved a perfect agreement. The proportions
of triples for which two annotators agree range from 87.07\% (highest)
to 74.06\% (lowest) and the agreement rate for four annotators was
even lower, 63.37\%. This indicates that matching synonyms with
definitions is a difficult task even for humans. On the other hand,
the reasonably high agreement rate suggests that the sample provides a
reasonable basis for evaluation. Accordingly we used the rating
produced by the first annotator of the pair with the highest agreement
as a baseline for our system.

\section{The basic procedure}
\label{sec:method}

Given a verb $V$, a synonym $Syn_V$ of that verb and a set of
definitions $D_V = \{d_1 \ldots d_n\}$ given for $V$ by the \tlfi ,
the task is to identify the definitions $d_i \in D_V$ of $V$ for which
$Syn_V$ is a synonym of $V$.

\paragraph{Mapping synonyms to definitions.} To assign a synonym $Syn_V$
to a definition $d_i$ of $V$, we proceed as follows: First we compare the index of the merged definitions
of $Syn_V$ with the index of each definition $d_i\in D_V$ using a
gloss-based similarity measure. Note that since
the intended meaning of the synonym is not given, we do not attempt
to identify it and use as the basis for comparison the union of the
definitions given by the \tlfi\ for each synonym. Next, the synonym $Syn_V$ is 
mapped onto the definition that gets the highest (non null) similarity score.

\paragraph{Evaluation.}
We evaluated the results obtained with respect to the reference sample
presented in the previous section as follows.

From the reference, we extracted the set of tuples $\langle$ V,
Syn$_V$, Def$_i \rangle$ such that Syn$_V$, is a synonym of V which is
associated with the definition Def$_i$ of V.

Recall is then the number of correct tuples produced by the system
divided by the total number of tuples contained in the
reference. Precision is the number of correct tuples produced by the
system divided by the total number of tuples produced by the system.

The baseline gives the results obtained when randomly assigning the
synonyms of a verb to its definitions.

\section{Experiments}
\label{sec:experiments}

\subsection{Comparing similarity measures}

To assess the impact of the similarity method used, we applied the 6
similarity measures listed in Table \ref{tab:sim-measures} namely,
simple word overlap, extended word overlap, extended word overlap
normalised, 1st order vectors and 2nd order vectors with and without a
tfidf threshold. These methods were implemented using Ted Pedersen's
Perl library \url{search.cpan.org/dist/WordNet-Similarity/} and
adapting it to fit our data\footnote{In particular, calls to the
  Princeton WordNet were removed.}.
\paragraph{Simple word overlap.} Simple word overlap between glosses
were introduced by \cite{lesk86} to perform word sense
disambiguation. The Lesk Algorithm which is used there, assigns a sense to a target word in
a given context by comparing the glosses of its various senses with
those of the other words in the context. That sense of the target word
whose gloss has the most words in common with the glosses of the
neighbouring words is chosen as its most appropriate sense. 

Similarly, here we use word overlap to assess the similarity between a
verb definition and the merged definitions of a synonym. Given a set
of verb definitions and a synonym, the synonym will be matched to the
definition(s) with which its definitions has the most words in common (and at least
one).
\paragraph{Extended word overlap.} 
The scoring mechanism of the original Lesk Algorithm does not differentiate between single word and phrasal overlaps. \cite{banerjeePedersen2003} modifies the Lesk method of comparison in two ways. First, the glosses used for comparison are
extended by those of related WordNet concepts and second, 
the scoring mechanism is modified to favour glosses containing phrasal
overlaps. An $n$ word overlap is assigned an $n^2$ score.
Because the French EuroWordNet is relatively
under-developed\footnote{The French EuroWordNet (\textsc{Frewn}) contains 3
  777 verbs. Since \cite{syndic5} alone lists 5 027 verbs, it is clear
  that a \textsc{Frewn} based extended gloss overlap measure would only
  partially be applicable.}  we did not modify the comparison to take
into account WordNet related glosses\footnote{As mentioned in the
  introduction \ref{sec:introduction}, an alternative WordNet for French is
  being developed by \cite{sagot08a}. It cannot be used to integrate
  in the comparison glosses of WordNet related words however because
  the glosses associated with synsets are the Princeton WordNet
  English glosses.}. We did however modify it to take into account
phrase overlaps using the same scoring mechanism as Banerjee and
Pedersen in \cite{banerjeePedersen2003}\footnote{Recall (cf. Section~\ref{sec:data}) that the index
  of a definition is the {\it list} of lemmas for the open class words
  occurring in that definition. The order in the list reflects the
  linear order of the corresponding words in the definition. 
}.
\paragraph{Extended word overlap normalised.} The extended word
overlap is normalised by the number of words occurring in the
definitions being compared.
\paragraph{First order vectors.}
A first order word vector for a given word indicates all the first
order co-occurrences of that word found in a given context (e.g., a
\tlfi\ definition). Similarity between words can then be computed
using some vector similarity measure. 
For each verb $V$, we build weighted word vectors for each of its definitions $d_V^i$
and for each of its synonyms. The dimensions of these vectors are the
lemmatised words occurring in the definitions of $V$ whose tf.idf is
different from 0. \footnote{The weights are computed as follows: For a definition $d_V^i$, the weight of each word
$w_j$ is the number of occurrences of $w_j$ in $d_V^i$ divided by
the number of occurrences of $w_j$ in all definitions of $V$. For a
synonym $Syn_V$, the weight of $w_j$ is the number of occurrences of
$w_j$ in the definitions of $Syn_V$ divided by the number of
occurrences of $w_j$ in all definitions of $V$.}
The similarity score between a verb definition $d_V^i$
and a synonym $Syn_V$ is the product of the two corresponding vectors.

\paragraph{Second order word vectors with and without tf.idf cutoff.} Second order vectors are
derived from first order vectors as follows. For each verb/synonym
definition, the corresponding second order vector is the sum of the
first order vectors\footnote{In contrast to the vectors used in the
first order approach, the dimensions of the first-order vectors used
to compute the 2nd order vectors are the lemmatised open class words
of all definitions in the \tlfi\ (not just the words occurring in
the definitions of a given verb).
} defined over the words occurring in this definition. The
second order vectors ``average'' the direction of a set of
vectors. If many of the words occurring in the definition have a
strong component in one of the dimensions, then this dimension will be
strong in the second order vector. In other words, the second order
vector helps pinning down the strength of the different dimensions in
a given definition.

The similarity score between a verb definition and a synonym is the
product of the two corresponding second order vectors. We compare two
versions of the second order word vectors approach, one where a tf.idf
cut-off is used to trim down the word space and another where it
isn't. 

The results obtained by the various measures are given in Table~\ref{tab:sim-measures} (left side). 
%

\begin{table}[htb] 
\small
\begin{center} 
  \begin{tabular}{|l|c|c|c||c|c|c|}
    \hline 
    \hline & \multicolumn{3}{|c||}{\textbf{No refl. dist.}} & \multicolumn{3}{|c|}{\textbf{With
        refl. dist.}} \\ \hline \textbf{Meas.} &
    \textbf{R} & \textbf{P} & \textbf{F} & \textbf{R} & \textbf{P} &
    \textbf{F} \\ 
    \hline 
    \hline 
    baseline & 0.45 & 0.32 & 0.38 & 0.44 & 0.43 & 0.44 \\ 
    \hline 
    \hline 
    Over 1 & 0.72 & 0.51 & 0.60 & 0.70 & 0.68 & 0.69\\ 
    \hline 
    Over 2 & 0.72 & 0.51 & 0.60 & 0.70 & 0.68 & 0.69 \\ 
    \hline
    Over 3 & \textbf{0.73} & \textbf{0.51} & \textbf{0.60} & \textbf{0.71} & 0.67 & \textbf{0.71} \\ 
    \hline 
    \hline
    WV 1 & 0.73 & 0.51 & 0.60 & 0.70 & 0.69 & 0.70\\ 
    \hline 
    WV 2 & 0.71 & 0.50 & 0.59 & 0.70 & 0.69 & 0.69\\ 
    \hline 
    WV 3 & 0.72  & 0.50 & 0.59 & 0.70 & \textbf{0.70} & 0.69\\ 
    \hline 
    \hline 
  \end{tabular} 
  \caption{\textbf{Precision, recall and F-measure for various similarity
    measures, with (right side) and without (left side) reflexive/non reflexive distinction.} The similarity measures are the following:
  \emph{Over 1}: Simple word overlap, \emph{Over 2}: Extended word overlap, \emph{Over 3}: Extended word overlap normalised, \emph{WV 1}: First order vectors, \emph{WV 2}: Second order vectors, without tf.idf cut-off, \emph{WV 3}: Second order vectors, with tf.idf cut-off. Best scores are set in bold face\protect\footnotemark.}
  \label{tab:sim-measures}
\end{center}
\end{table}
\footnotetext{Please note that the values shown here have been computed with higher precision and then rounded, therefore some differences in scores may no longer be visible.}
A first observation is that our synonym-to-definition mapping
procedure systematically outperforms the random assignment
baseline. Thus, despite the brevity of dictionary definitions, gloss
based similarity measures appear to be reasonably effective in
associating a synonym with a definition on the basis of its own
definitions. 

A second observation is that no similarity measure clearly yields
better results than the others. This suggests that word overlap
between \tlfi\ definitions is a richer source of information for
synonym sense disambiguation (SSD) than other more indirect contextual
cues such as the distributional similarity of the words occurring in
the definitions (first order word vector approach) or of the words
defined by the words occurring in the definitions (second order word
vector approach).

\subsection{Linguistic preprocessing}
\label{subsec:refl}

A single \tlfi\ verb entry might encompass several very different
uses/meanings of this verb. Typically, it might include definitions
that relate to the reflexive use of that verb, to a non reflexive use
and/or to collocational use. 

The approach presented in the previous section does not take such
distinctions into account and is therefore prone to compare apples and
oranges. It will for instance select the synonyms of a verb $V$ and
match these into all its definitions independent of whether these
definitions reflect a reflexive or a non reflexive usage. This is
clearly incorrect because the synonyms of a verb $V$ are not
necessarily synonyms of its reflexive form. For example, the synonyms
of the non reflexive form {\it abandonner} (to abandon) listed in Fig.~\ref{fig:abandonner} are clearly distinct from those of the reflexive
form {\it s'abandonner} (to give way).

\begin{figure}
  \centering
  \small
  \begin{tabular}{p{7.5cm}}
    {\sc Abandonner}: (1) se dessaisir, renoncer \`a, se d\'eposs\'eder, se d\'epouiller, abdiquer, se d\'emettre, d\'emissionner, se d\'esister, r\'esigner, renoncer \`a, sacrifier, c\'eder, confier, donner, l\'eguer (2) conc\'eder, accorder (3) exposer (ancient), d\'elaisser, l\^acher, tomber, larguer (fam.), plaquer (fam.) \ldots\\
    {\sc S'abandonner}: se livrer, succomber, c\'eder, se donner, s'\'epancher, se fier, se reposer sur
  \end{tabular}
  \caption{Sample (reflexive and non-reflexive) synonym dictionary entry of \emph{(s') abandonner}, (\emph{to abandon}).}
  \label{fig:abandonner}
\end{figure}

Hence matching e.g., the
synonyms of {\it abandonner} onto definitions corresponding to a
reflexive use of the verb will result in incorrect synonym/definition
associations.


To account for these observations, we developed an approach that aims
to take into account the reflexive/non reflexive distinction. The
approach differs from the procedure described in the previous section
as follows: First, we automatically differentiated both in the handbuilt reference
and in the automatically extracted verb entries between the
reflexive and the non reflexive usage of a verb. For each verb with
the two types of usage, we constructed two entries each with the
appropriate definitions. The synonym selection is then done with respect to a verb entry i.e.,
with respect to either a reflexive or a non reflexive usage. 
%

As a result, similarity measures were applied between the
definitions of verbs corresponding to the same type of usage. In other
words, the definitions of a synonym associated with a given verb usage
(reflexive {\it vs.} non-reflexive) were compared only with
the definitions of this particular usage.

The results obtained on the basis of this modified procedure are given
in Table~\ref{tab:sim-measures}, right side.
Unsurprisingly while precision increases, recall decreases. 
The increase in precision indicates that this linguistically more
constrained approach does indeed support a better matching between
synonyms and definitions. The decrease in recall can be explained by
several factors. First, the information contained in the
\tlfi\ concerning reflexive and non reflexive usage is irregular so
that it is sometimes difficult to automatically distinguish between
the definition of a reflexive usage and that of a non reflexive
usage. Second, the synonym dictionary might fail to provide synonyms
for a reflexive usage listed by the \tlfi . Third, a reflexive verb
listed in the synonym dictionary might fail to have a corresponding
entry (and hence definition) in the \tlfi . All of these cases
introduce discrepancies between the reference and the system results
thereby negatively impacting recall.  

In short, while a finer linguistic processing of the data contained in
the \tlfi\ might help improve precision, a better recall would involve
enriching both the synonym and the \tlfi\ dictionaries. 


\section{Related work}
\label{sec:related}

Our work has connections to several research areas namely, word sense
disambiguation (we aim to identify the meaning of a synonym and more
specifically, to map a synonym to one or more dictionary definitions
associated by a dictionary with the verb of which it is a synonym),
synonym lexicon acquisition (we plan to use the method presented here
to merge the five synonym lexicons into one) and WordNet construction
(by identifying sense based synonym sets i.e., synsets).

\paragraph{Word sense disambiguation (WSD)} uses 
four main types of approaches namely, lexical knowledge-based methods
which rely primarily on dictionaries, thesauri, and lexical knowledge
bases \cite{lesk86,klbasedWSD}, without using any corpus evidence;
supervised and semi-supervised approaches \cite{supervisedwsd} which
make use of sense annotated data to train or start from and
unsupervised methods \cite{unsupervisedwsd} .

The approach presented here squarely fits within the lexical
knowledge-based methods in that it exclusively uses dictionary
definitions to disambiguate words. Supervised and semi-supervised
approaches were not considered because of the absence of sense
annotated data for French. Moreover, as shown by the construction of
the reference sample and the agreement rate obtained 
(cf. Section~\ref{sec:data}), the fact that we are working on
disambiguating synonyms (as opposed to a set of arbitrary words) out
of context makes sense annotation a lot more difficult than for the
standard WSD task.

It would in principle be possible to use an unsupervised approach and
attempt to disambiguate synonyms on the basis of raw 
corpora. Such approaches however are not based on a fixed list of
senses where the senses for a target word are a closed list coming
from a dictionary. Instead they induce word senses directly from the
corpus by using clustering techniques, which group together similar
examples. To associate synonyms with definitions, it would therefore be
necessary to define an additional mapping between corpus induced word
senses and dictionary definitions. As noted in \cite{agirreEtAl},
such a mapping usually introduces noise and information loss however. 

\paragraph{Synonym lexicon construction.} As noted above and further discussed in
Section~\ref{sec:conclusion}, the method described in this paper can
be used to merge the five synonym dictionaries mentioned in section
\ref{sec:data} into a single one. In this sense, it is related to work
on synonym lexicon construction. Much work has recently focused on
extracting synonyms from dictionaries and/or from corpora to build
synonym lexicons or thesauri. Thus,
\cite{hindl90,crouchYand92,grefensette94} extract synonyms from large
monolingual corpora based on the idea that similar words occur in
similar context; \cite{barzilayMckeown01} used a bilingual
corpus; \cite{sennelart02} use the structure of monolingual
dictionaries; and \cite{wuZhou03} combine both monolingual and
bilingual resources. Such approaches are fundamentally different from
the work presented here in two main ways. First, they aim to extract
synonyms from linguistic data and thereby often yield
``associative'' lexicons rather than synonymic ones. In other words,
these approaches yield lexicons which often associate with a word,
synonyms but also antonyms, hypernyms or simply words that belong to
the same semantic field. In contrast, we work on a predefined base of
synonyms and the lexicon we produce is therefore a purely synonymic
lexicon. Second, whereas we associate synonyms with a predefined list
of senses, existing work on synonym lexicon construction usually
doesn't and is restricted to identifying sets of synonyms (or
semantically related words).

\paragraph{WordNet and thesaurus construction.}
Grouping synonyms in sets reflecting their possible senses effectively
boils down to identifying synsets i.e., sets of words having a common
meaning. In this sense, our work has some connections with work on
WordNet development and more precisely, with a {\it merge} approach to
WordNet development that is, with an approach that aims to first
create a WordNet for a given language and then map it to existing
WordNets.
Recently, \cite{sagot08a,sagot08b} have
presented an {\it extend} approach to WordNet construction for French based
on a parallel corpus for 5 languages (French, English, Romanian,
Czech, Bulgarian). Briefly the approach consists in first extracting a
multilingual lexicon from the aligned parallel corpora and second, in
using the Balkanet WordNets to disambiguate polysemous words. The
approach relies on the fact that the WordNets for English, Romanian,
Czech and Bulgarian all use the same synset identifiers. First, the
synset identifiers of the translations of the French words are
gathered. Second, the synset identifier shared by all translations is
assigned the French word. In this way, and using various other
techniques and resources to assign a synset identifier to monosemous
words, \cite{sagot08a,sagot08b} produces a WordNet for French called
WOLF (freely available WordNet for French) that replicates the
Princeton WordNet structure.

Like work on synonym extraction, the WOLF approach differs from ours
in that synonyms are automatically extracted from linguistic data
(i.e., a parallel corpus and the Balkanet WordNets) rather than taken
from a set of existing synonym dictionaries thereby introducing
errors in the synsets. \cite{sagot08a,sagot08b} report a precision of
63.2\% for verbs with respect to the French EuroWordNet. A second
difference is that our approach associates synsets with a French
definition (from the \tlfi ) rather than an English one (from the
Princeton WordNet via the synset identifier). A third difference is
that we do not map definitions to a Princeton WordNet synset
identifier and therefore cannot reconstruct a network of lexical
relations between synsets. More generally, the two approaches
are complementary in that ours provides the seeds for a {\it merge}
construction of a French WordNet whilst \cite{sagot08a,sagot08b}
pursue an extend approach.  

\section{Conclusion and future work}
\label{sec:conclusion}

We have presented an automatic method for assigning synonyms to definitions with a reasonably high 
F-score of at best, 0.70 (P=0.67,R=0.71). Future work will focus
on two main points.

First, we will explore ways of improving these results. In particular,
we will investigate in how much the structure of a dictionary entry
can be used to enrich a definition. As mentioned in Section~\ref{sec:data}, a dictionary entry has a hierarchical structure which
is often used by the lexicographer to omit information in definitions
occurring lower down in the hierarchy. Automatically enriching the
\tlfi\ definitions by inheriting information from higher up in the
dictionary entry might result in definitions which, because they
contain more information, provide a better basis for similarity
measures. Similarly to the distinguishing treatment of reflexive/non
reflexive usages discussed in section \ref{subsec:refl}, we will also develop
a separate treatment of definitions involving verbal collocations (as
opposed to isolated verbs).


Second, we will use this method to merge the synonym dictionaries
into one where each word is associated with a set of (\tlfi )
definitions and each definition with a set of synonyms. We will then
investigate, on the basis of the resulting merged synonym dictionary,
how to reconstruct the lexical relation links used in WordNet. To
this end, we intend to explore in how far translation and ontology
enrichment techniques \cite{bruinEtAL04} can be applied
to enrich our synonym lexicon and align it with the Princeton
WordNet. In this way, we can build on the WordNet structure given
by the Princeton WordNet and enrich the synsets derived from the five
synonym dictionaries with translations of the related English
synonyms.


\bibliographystyle{abbrv}  
\begin{scriptsize}
\bibliography{syn2-ranlp09}  
\end{scriptsize}

\end{document}